\documentclass[letterpaper, 10 pt, conference]{ieeeconf}  

\IEEEoverridecommandlockouts                              
\usepackage{graphicx}
\usepackage{tabularx}
\usepackage{amsmath,amssymb} 
\usepackage{color}
\usepackage{authblk}
\usepackage[%
    colorlinks=true,
    pdfborder={0 0 0},
    linkcolor=red
]{hyperref}
\usepackage[maxbibnames=99]{biblatex}
\usepackage{amsmath}
\usepackage{booktabs}
\usepackage{arydshln} 
\title{\LARGE \bf
BEVLoc: Cross-View Localization and Matching via Birds-Eye-View Synthesis
\thanks{$^*$ C. Klammer and M. Kaess are part of the Robotics Institute, Carnegie Mellon University, Pittsburgh, PA, 15213 \authorcr Email: {\tt \{cklammer, kaess\}@cs.cmu.edu}}
\thanks{$^\dagger$ C. Klammer is part of KEF Robotics, Pittsburgh, PA, 15206 \authorcr Email: {\tt chris@kefrobotics.com}}
}

\author{Christopher Klammer$^*$$^\dagger$ \\ Carnegie Mellon University \and Michael Kaess$^*$ \\ Carnegie Mellon University}

\begin{document}

\maketitle
\thispagestyle{empty}
\pagestyle{empty}

\begin{abstract}

Ground to aerial matching is a crucial and challenging task in outdoor robotics, particularly when GPS is absent or unreliable. Structures like buildings or large dense forests create interference, requiring GNSS replacements for global positioning estimates. The true difficulty lies in reconciling the perspective difference between the ground and air images for acceptable localization.

Taking inspiration from the autonomous driving community, we propose a novel framework for synthesizing a birds-eye-view (BEV) scene representation to match and localize against an aerial map in off-road environments. We leverage contrastive learning with domain specific hard negative mining to train a network to learn similar representations between the synthesized BEV and the aerial map.

During inference, BEVLoc guides the identification of the most probable locations within the aerial map through a coarse-to-fine matching strategy. Our results demonstrate promising initial outcomes in extremely difficult forest environments with limited semantic diversity. We analyze our model's performance for coarse and fine matching, assessing both the raw matching capability of our model and its performance as a GNSS replacement.

Our work delves into off-road map localization while establishing a foundational baseline for future developments in localization. Our code is available at: \href{https://github.com/rpl-cmu/bevloc}{https://github.com/rpl-cmu/bevloc}
\end{abstract}

\section{INTRODUCTION}

In the realm of navigation, the presence of reliable and consistent GPS is paramount in existing localization solutions to help constrain the pose optimization to mitigate drift. However, GPS often is disrupted by interference from excessive forestry, overarching structures, atmospheric conditions, or adversaries intentionally jamming GPS signals. When GPS is absent, visual odometry (VO) and visual-inertial odometry (VIO) systems will eventually drift and become globally inaccurate without any sort of global reference or registration \cite{monocular_vio_comparison} \cite{vio_concise_review}. These inaccurate estimates induce risks such as entering intraversable areas, diverging from the planned path, and damage to the environment or the robot itself.

\begin{figure}[!ht]
    \centering
    \includegraphics[width=0.9\linewidth]{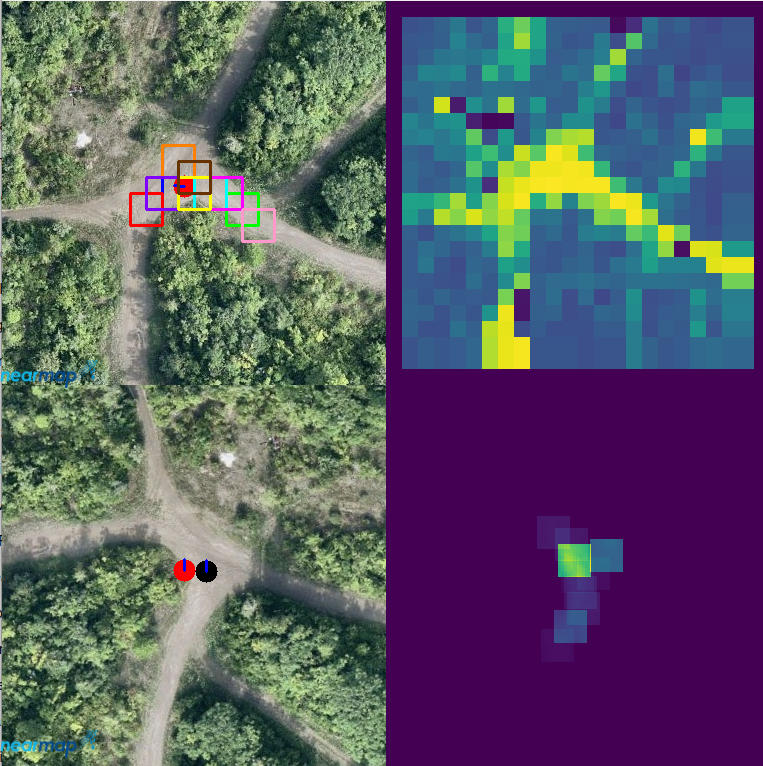}
    \caption{Visualization of the coarse to fine matching strategy used by BEVLoc. The top panes show the coarse matches with the map aligned image and the corresponding correlation volume. The bottom panes show the map rotated by the predicted yaw angle and the refinement of the coarse location to create a probabilistic prediction for the localization by weighting the correlation maps over the top $k$ coarse matches. The red circle denotes the ground truth localization, the black circle denotes the predicted localization.}
    \label{fig:VizMatching}
\end{figure}

The challenges and difficulties of state estimation are only exacerbated by using only vision sensors without the use of a reliable GPS signal. Though deploying high cost LiDAR sensors may delay the presence of drift, it is still inevitable without re-visitation or loop closures \cite{legoloam2018shan}\cite{Zhang-2014-7903}. To combat this, we explore using a prior map of the environment to aid our GPS-denied localization pipeline by visually matching against and attempting to re-localize within the known map.

While many existing methods focus on rich semantic content present in on-road datasets, our focus rests in scenarios with sparse semantic information. The main challenge lies in harvesting whatever information possible to create geometric and semantic features to match against. However, this comes with challenges in the form of consistency between the aerial map and first-person-view (FPV) images - seasonal changes and change in vegetation can add additional complexity.

We seek to solve the "kidnapped robot problem" for an autonomous ground robot given only a map of the environment, a prior GPS location, and the vision sensors onboard the robot. Our motivation lies in focusing on localization in unstructured and off-road environments. Our work proposes a vision-centric pipeline to learn similar representations between aerial maps and ground camera images in order to localize the robot within the map. Using the semantics and geometry of the scene, our method  creates a synthetic birds-eye-view (BEV) representation to reconcile the perspective difference between aerial and the ground images. 

Our contributions include a contrastive learning framework to provide an embedding for aerial and ground images to be used for downstream matching. We propose a coarse-to-fine matching approach to fuse a registration estimate with the sensor data in order to localize within an aerial map.

\begin{figure}[!t]
    \centering
    \begin{minipage}{0.48\linewidth} 
        \centering
        \includegraphics[width=\linewidth]{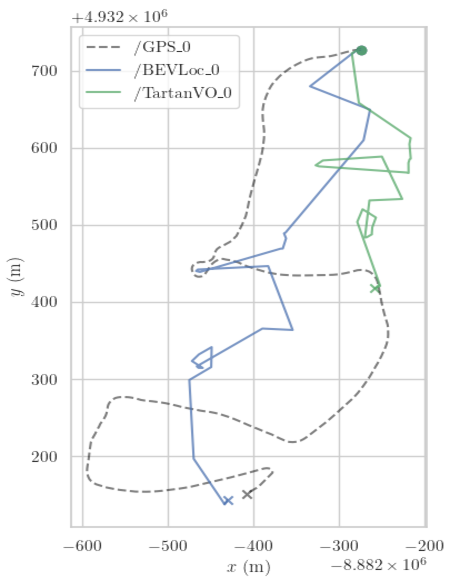} 
        \label{fig:Traj2}
    \end{minipage}
    \hspace{0.0025\linewidth} 
    \begin{minipage}{0.48\linewidth} 
        \centering
        \includegraphics[width=\linewidth]{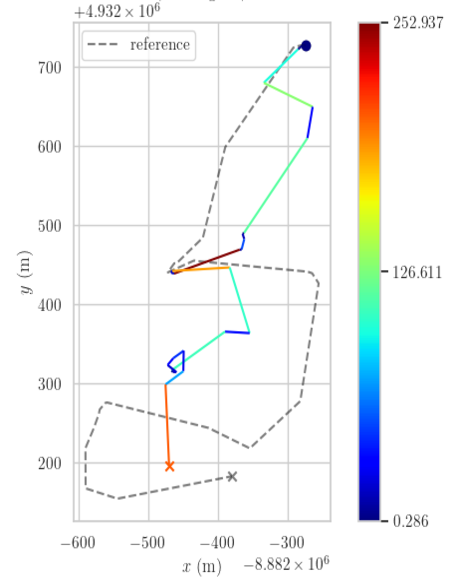} 
        \label{fig:RPEPlot}
    \end{minipage}
    \caption{Illustration of our method's performance, leveraging registration estimates to mitigate drift from visual odometry. \textbf{Left:} Comparison of BEVLoc and Tartan VO against the ground truth GPS trajectory. \textbf{Right:} RPE performance of BEVLoc throughout a long trajectory against GPS ground truth.}
    \label{fig:Traj}
\end{figure}

\section{Related Works}
Ground-to-air matching is not a new task, oftentimes, existing works have treated this as an image retrieval problem for georegistration. Many existing works have looked at cross-view localization as an image retrieval problem for geolocalization. These methods showed initial promise, using generative models to bridge the gap between ground and aerial imagery to help learn more robust feature descriptors while utilizing contrastive learning to further close the domain gap \cite{regmi_bridging_2019}. Other works have focused primarily on cross-view correspondence, encoding semantic correspondences over time to be trained in a contrastive manner against reference satellite imagery embeddings \cite{10.1007/978-3-031-26319-4_8}.

Additionally, recent works have explored the potential of lifting sensor information from the ground for perception. Initial works looked at lifting multi-view camera images to a shared birds-eye-view representation (BEV), maintaining visual features from cameras and learn depth priors \cite{hu_fiery_2021} \cite{philion_2020_lift}.  Often, works combine these visual features with geometric features from LiDAR and/or radar in order to maximize perception accuracy for downstream birds-eye-view feature maps in the same detection and segmentation tasks. These methods use LiDAR to model the occupancy of the scene, creating geometric features and using splatting operators or pure projection of feature maps to create a shared representation \cite{can_understanding_2022} \cite{pointpillars} \cite{li2022bevformer} \cite{liu_bevfusion_2022} \cite{Yang2022BEVFormerVA}.  Notably, SimpleBEV \cite{harley2022simple} took a closer look, analyzing what strategies made the biggest difference for perception tasks, highlighting the utility of bilinear interpolation along rays instead of predicting monocular depth priors. These methods focus on perception in a fixed three-dimensional grid around the robot ego-frame and infer the localization of objects rather than the robot itself. We take inspiration from these works to lift image features to a birds-eye-view, using cameras available onboard and relying on stereo depths as a depth prior to better constrain the geometry of the scene.

Most similar to our work, \cite{sarlin2023snap} \cite{sarlin2023orienternet} have looked at 3DoF localization given multi-view images at an instantaneous point in time for widely available public street maps. These works focus on the strength of the semantics for localization and learning depth priors for neural maps. Their pose alignment strategy demonstrates the potential of BEV featurization and its accuracy in 3DoF visual positioning and semantic mapping given only a single image. As well, they look towards creating a contrastive learning framework as well, using pose supervision and mining high likelihood false positives through RANSAC adjacent strategies. On top of this and focusing on extremely hard negative samples, \cite{robinson_contrastive_2021} provides well suited guidance on mining said negatives for contrastive learning frameworks to create representations for our usage.

Lastly, visual place recognition has received a resurgence with the introduction of foundational models \cite{oquab2023dinov2}. Their matching capability has been demonstrated for visual place recognition and has been shown to work across a diverse set of scenes \cite{NetVLAD} \cite{keetha_anyloc_2023}. More recent works have shown the utilization of visual place recognition on a real aerial robotics platform on real datasets \cite{10.1007/978-3-031-26319-4_8}. This furthers our confidence in integrating foundational models for matching on real robotics platform and on state estimation pipelines.
\section{Methodology}

\begin{figure*}[!t]
    \centering
    \includegraphics[width=\textwidth]{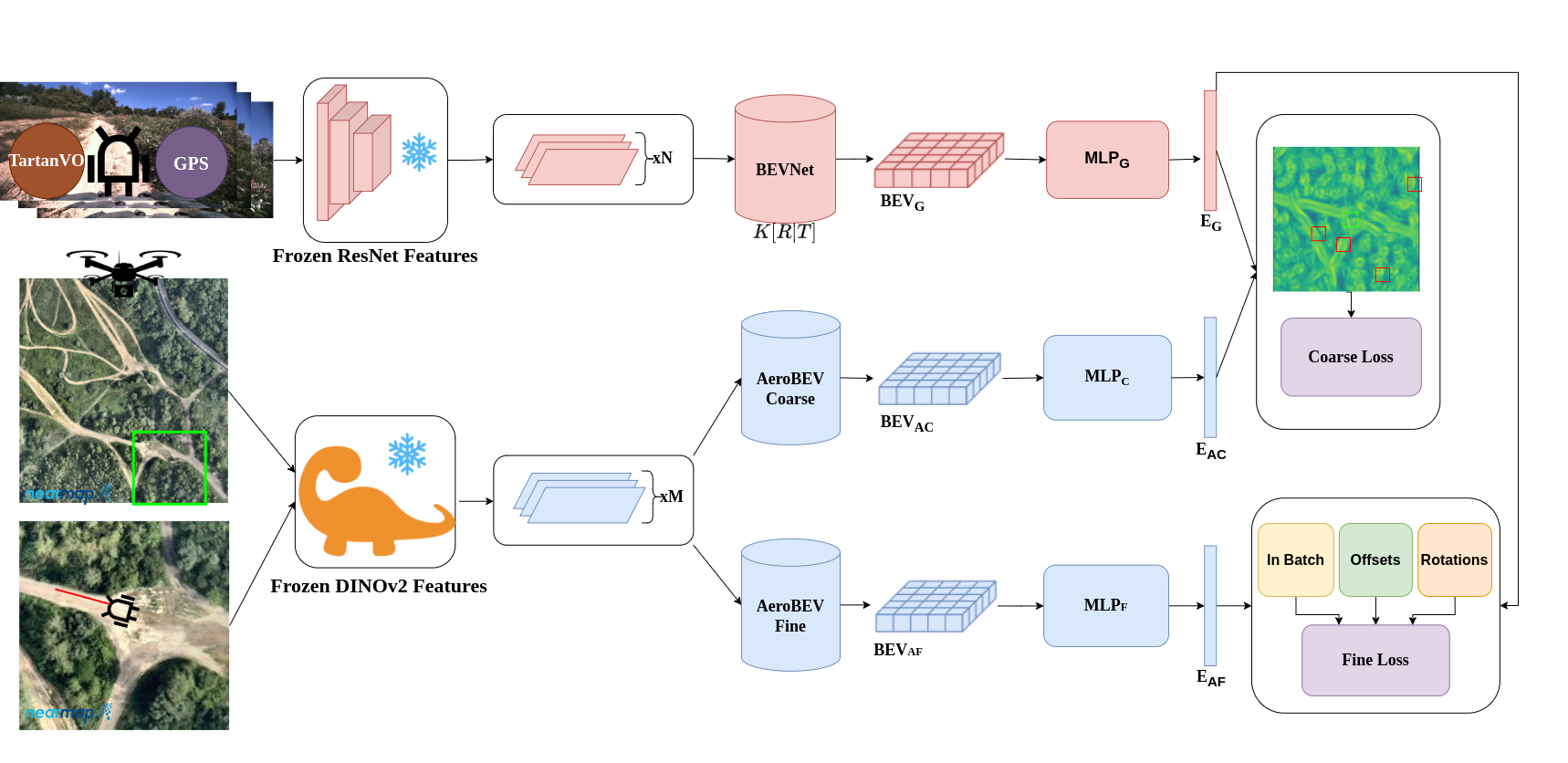}
    \caption{BevLoc Contrastive Learning Training Pipeline. Feature maps are encoded from ground and aerial camera images. The ground features are lifted to 3D to create a semantic and temporally consistent BEV feature map to be compared against the aerial feature map. The embeddings are created and used to find hard negatives near the prior location and prior rotation to learn how to match ground to aerial images.}
    \label{fig:contrastive-learning-pipeline}
\end{figure*}

\subsection{Problem Setup}
Given a prior 3DoF unmanned ground vehicle (UGV) pose $P_0 = (x, y, \theta_{yaw})$, our method seeks to infer a global state estimate using vision sensors alone. We utilize the Tartan Drive 2.0 dataset \cite{sivaprakasam_tartandrive_2024}\cite{triest2022_tartandrive} for all of the real-world UGV data. This includes a Carnegie Robotics MultiSense S21 sensor with FPV images and depth at 10Hz. From these sensors, we aim to create a synthesized BEV feature map $BEV_G$ can be used to match against cropped regions of the aerial maps, encoded as feature maps $BEV_A$ . We utilize TartanVO\cite{tartanvo2020corl} to estimate local odometry $(\Delta x, \Delta y, \Delta \theta)$ between consecutive frames capture local motion of the robot, aiding in building a temporally informed representation for our method. Ultimately, we seek to match a ground representation $BEV_G$ against multiple "aerial crops" or cut out locations centered at a GPS location. These representations, denoted as $BEV_A$, provide a global state estimate to re-localize the robot within the map and to correct for drift in the robot's trajectory. As for assumptions:
\begin{enumerate}
    \item We assume the sensors are calibrated and the intrinsics and extrinsics are all known.
    \item We assume the map is of known resolution and orthorectified.
    \item We are given purely the map images and the sensors on-board the UGV. There are no known semantic images to simplify the learning process.
\end{enumerate}

\subsection{Lifting Features from the Ground}
Our first major task is to lift camera images on the ground to compare against semantics in the aerial image. The top branch of \ref{fig:contrastive-learning-pipeline} shows the process of processing camera images to be lifted into a BEV feature representation. For this encoding, we elect to use a ResNet-101 convolutional network \cite{he_deep_2015} over foundational models. The reason for this lies in the stride size of 14 in the foundational backbones creates much lower resolution feature maps compared to ResNet-101, which could spawn ambiguities about the spatial locality of features. In turn, we theorize this could potentially decrease the localization accuracy of the features when projecting them to the feature volume.  

We choose to place pixels of 2D feature maps into a 3D feature volume using the corresponding depth image to capture the visual features in a discretized grid around the robot. These features are placed at location $X_{feat}$ in the corresponding voxel grid with voxel size of $[V_x, V_y, V_z] = [.3m, .3m, 3.0m]$ and grid size of $[G_x, G_y, G_z] = [32m, 32m, 16m]$. The transformation of features from camera to the grid, unprojecting the features followed by a translation to a grid frame such that the robot is in the middle of the grid as such that:
\begin{equation}
    X^{feat} = T^{grid}_{ned}T^{ned}_{cam} \Pi(
\begin{bmatrix}
    xz\\
    yz\\
    z
\end{bmatrix})
\end{equation}
This feature volume accumulates features, which are then averaged using the cumsum trick \cite{philion_2020_lift} to efficiently summarize features per voxel. Next, we elect to use max pooling over the pillars like \cite{sarlin2023snap}\cite{harley2022simple} to create a geometrically consistent BEV feature map $BEV^{(i)}_G$ for a given timestep $i$ and highlight important semantics the map would reflect. We elect to use temporal information by utilizing the local odometry estimates to build a locate feature map and concatenate many $BEV_G$ maps for a more rich semantic profile. We do this by using the last pose in the batch as the reference pose  $T = [R|t] = [I|0]$ and each antecedent pose relative to the reference. A pose at time $k$ is calculated as:
\begin{equation}
   T^{odom}_k = \Pi^k_{i = ref} T_i^{-1} 
\end{equation}
The final BEV feature map at a timestep $k$ includes combining the features at every pixel, placing them into the grid then transforming the grid to respect the odometry.
\begin{equation}
   BEV_k = T^{odom}_k X^{feat}_k 
\end{equation}
The final map $BEV_G$ concatenates the BEV feature maps for each time step channel-wise for a more semantically rich and temporally sensitive representation for the entire batch of size $B$, denoted by the concatenation operator $\bigoplus$.
\begin{equation}
    BEV_G = \bigoplus^{B} BEV_k
\end{equation}
Our method, much like other methods, \cite{harley2022simple}\cite{hu_fiery_2021}\cite{philion_2020_lift} send this birds-eye feature map through a BEV encoder to compress and summarize the features for each spatial cell before passing through a linear layer to create the embedding $E_G$ which is used for a representation to be compared against the aerial map.

\subsection{Encoding the Aerial Images}

\subsubsection{Coarse Network}

We train an AeroBEV coarse network using no prior rotation to receive a general localization for the robot. We use a correlation mining strategy to help select hard negatives, discussed more in \ref{sec:HardNegativeCoarse}

$E_{AC}$ is produced by taking all of the map aligned cells and encoding them with a DINOv2 encoder. Since we look to capture coarse features rather than fine ones, DINOv2 is a strong backbone candidate for capturing the general semantics of the map cell and will likely be less prone to small discrepancies or noise in the map cell. This network is fairly simple, focusing on taking the feature map and feeding forward to an embedding using a linear layer, creating an embedding for each grid cell. The aim is to provide a descriptive embedding that can be compared to the sequence of recent ground images.

\subsubsection{Fine Network}
We train a separate AeroBEV network, denoted as AeroBEVFine, using the robot's prior rotation and aligning the aerial crop to be rotated around the robot's current position. Positives are denoted as high correlation matches with a rotation within $\tau_{\theta} = 10$ degrees of the robot and within $\tau_{far} = 3$ meters of the robot. We have separate mining modules to mine map locations are responsible for ensuring this network is neither rotation invariant nor translation invariant \ref{sec:HardNegativeFine}. This network is identical to that of the coarse network besides the loss function which we discuss below.

$E_{AF}$ is produced by taking all of the robot-aligned cells and encoding them with a DINOv2 or ResNet-101 encoder. That is, we use the current estimate $\theta_{yaw}$ to rotate the map and create robot-aligned embeddings. This network architecture is the exact same as the coarse network besides the encoder that is used. We select to use a smaller DINO-B encoder or ResNet-101 encoder to reduce the size of the features. Notably,  we found that the ResNet-101 encoder received superior results to the foundational models, despite the strength in expressing scene semantics, we discuss this in more detail in \ref{ExperimentalResults}.

\subsection{Map Matching as a Contrastive Learning Problem}
Given an embedding $E_G$ for the ground images and an embedding $E_A$ for the aerial crops centered around a GPS location, we now aim to minimize the distance between the embeddings for the last $n$ camera images and the aerial crop at the given GPS location and yaw angle. 

\subsubsection{General Loss}
We use a generalized cosine loss with a margin $m$ which specifies the distance in the embedding space we look to have between positive samples and negative samples.

\begin{equation}
    \mathcal{L_+}(..., y) = \frac{1}{|\mathcal{E}_A+|} \sum_{E_A \in \mathcal{E}_A+} y \cdot d(E_G, E_A)
\end{equation}
\begin{equation}
    \mathcal{L_-}(..., y, m) = \frac{1}{|\mathcal{E}_A-|} \sum_{E_A \in \mathcal{E}_A-} (1 - y) \cdot(m - d(E_G, E_A))
\end{equation}

\subsubsection{Mining Negative Coarse Aerial Embeddings}\label{sec:HardNegativeCoarse}
We define the coarse problem over the aerial crop of the map around the ground truth GPS location. We extract a local map of size $[A_x, A_y] = [384 pix, 384 pix]$ and utilize the set of all non-intersecting aerial embeddings $\mathcal{E}_A$ as our samples for the contrastive learning problem. During inference, this formulation changes slightly, with the most recently inferred pose of the robot being the center of the local map.

We define each region that creates an aerial embedding as the size of the ground robot's local map $[G_x, G_y]$. At training time, we choose only one positive to be the map cell in which the robot is closest to the center with the rest being negatives. With only one positive, we aim to use the general semantics in the cell to learn coarse localization predictions to be refined in the subsequent modules.

When mining negatives, we take each coarse embedding at the current time step and find the correlation with the ground embedding by taking the dot product. We mine negatives as any sample with: $corr(E_G, E_A) \ge 0$. As negatives become more difficult to mine, we establish an additional criteria that search for the top 25 percent of false positive correlation values and mark those as negatives. Therefore, we will always have, at least, 25 percent of the current map cells as negatives. We choose to not pass in all of the map cells to focus on learning useful areas to disassociate from the ground embedding while not hampering the network by overfitting on easy examples.

\subsubsection{Coarse Loss}
We define the coarse loss over the set of all non-intersecting aerial embeddings $\mathcal{E}_A$ and calculate their distance from the ground embeddings. We penalize the positive sample by its distance from the ground embedding. Meanwhile, we enforce a constant margin $m_c$ as a target distance for negative samples, encouraging negatives to be pushed as far away as possible. Empirically, we set $m_c = 1.5$.
\begin{equation}
    \mathcal{L}_{c} = \mathcal{L_+}(y) + \mathcal{L_-}(..., y, m_c)
\end{equation}
\subsubsection{Mining Negative Fine Aerial Embeddings}\label{sec:HardNegativeFine}
Our goal in the fine problem is to refine the coarse matches in order to receive high quality global localization estimates. Our method method samples offsets in a random direction with offset value samples from a uniform distribution, making it equally likely we will sample a positive or a negative near the ground truth location. We take each of these samples and accept them as positives if they are within a threshold $\tau_d$.
\begin{equation}
    x_o, y_o \sim U[0, \sqrt{2}\cdot \tau_d]
\end{equation}
\begin{equation}
   y_A = ||\begin{bmatrix}
    x_o & y_o
\end{bmatrix}||_2 < \tau_d 
\end{equation}
Likewise, we define rotations that are greater than 10 degrees to be negatives, taking care to handle angle wraparound. To mine the rotation negatives, we sample $n_{rot}$ angles and take their chips at the ground truth location to produce embeddings that are not invariant to fine rotations.

\subsubsection{Fine Loss}
For the fine loss, we have three separate components:

\begin{enumerate}
    \item An offset loss, which penalizes nearby negatives and penalizes hard positives by a Gaussian weight proportional to the physical distance from the ground truth location.
    \item A rotation loss, which discourages overly rotated aerial crops at the given location.
    \item A within-batch loss, which assigns poses within the batch positive or negative relative to the last pose
\end{enumerate}

When performing fine matching, it is desirable for our embeddings to have a sense of distance when the robot is at different rotations. To reflect this, we introduce a loss that pushes away map embeddings that are at acceptable localization but have an inaccurate heading for the robot. $m_R = 0.8$
\begin{equation}
  \mathcal{L}_R = \mathcal{L}(..., y_R, m_R)  
\end{equation}
For the offset aerial embeddings, we introduce additional Gaussian penalty loss $\mathcal{L}_G$ that artificially pushes away samples more the further they are from the ground truth location within the map
\begin{equation}
    \mathcal{G}(d, \sigma) = e^{-\frac{1}{2}\cdot (\frac{d}{\sigma})^2}
\end{equation}
We use RANSAC to sample a set number of positive and negative samples around the ground truth location. The final loss calculates a Gaussian weighted loss that takes into account the similarity to their embedding in addition to the physical distance $d_p$ away from the ground truth location. We set $\sigma_o = 64$ pixels and $m_o = 1.5$
\begin{equation}
    \mathcal{L}_O = \mathcal{G}(d_p, \sigma_o) \mathcal{L}(..., y_o, m_o)
\end{equation}
The within-batch loss checks for samples that are within a certain distance threshold $\tau_{far}$. In our 
experiments, $\tau_{far}$ is 3 meters and $m_B$ is 1.25.

\begin{equation}
    y_B^{(i)} = d(X_i, X_{ref}) <= \tau_{far}
\end{equation}
\begin{equation}
    \mathcal{L}_B = \mathcal{L}(..., y_B, m_B)
\end{equation}

\subsection{Coarse to Fine Matching} \label{Coarse To Fine Matching}
During the inference phase, we employ a heuristic coarse-to-fine matching strategy to harness the fine network. This strategy consists of:
\begin{enumerate}
    \item Identification of the top $k$ highly correlated matches and generation of a local correlation volume $C_{\theta}$ for a given yaw angle $\theta$.
    \item Outlier rejection by retaining high correlation estimates and employing multiple different orientations of the aerial crop to eliminate outliers.
\end{enumerate}

To construct $C_{\theta}$ we scan match around the coarse match for $x\in [-\frac{G_x}{2}, \frac{G_x}{2}, s_x],  y \in [\frac{-G_y}{2}, \frac{G_y}{2}, s_y]$ where $s_x = s_y = 1$ meter to find the highest correlation estimates. We weight these $k$ correlation volumes, choosing to weight the top 3 coarse matches' contribution to be 70 percent of the correlation volume.  Our assumption here is during our training process, we learned to lower the correlation values of nearby locations and will have a defined peak for the most probable estimate. However, this is very prone to outliers as some regions may look extremely similar to other regions in the map. Therefore, we propose looking at multiple orientations of map locations, targeting high correlation gradients across different angles. Notably, our work selects $\theta \in [-20, 20, s_{\theta}]$ with the positive being $\theta = 0$ and the other orientations being negatives.

We use the final fused correlation maps as our measurement covariance for the state estimate. Noise in the coarse and fine matching will be reflected in this covariance and is calculated by the outer product of the distance from the predicted location $\mu$ to the query location $x_i$, and weighted by a normalized probability for the cell $p_i$
\begin{equation}
    \Sigma_r = \sum_i p_i (x_i - \mu)(x_i - \mu)^T
\end{equation}
\begin{figure}[!t]
    \centering
    \includegraphics[width=0.9\linewidth]{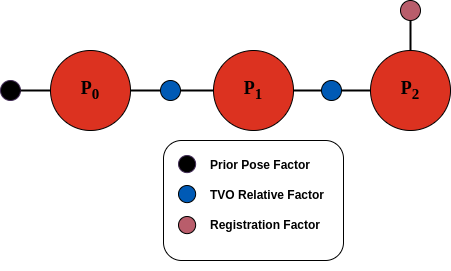}
    \caption{An illustration of the global pose graph which corresponds to the 3DoF pose of the vehicle. As a vision-only pipeline, we utilize relative visual odometry measurements and high quality registration estimates.}
    \label{fig:FactorGraph}
\end{figure}
\subsection{Pose Estimation via Non-Linear Optimization}
We employ a framework for non-linear pose optimization using a factor graph \cite{KaessFactorGraphs}. This approach allows us to refine estimates from multiple streams of information.

Our factor graph looks to solve a non-linear least squares optimization where we estimate all poses up to a time $t$ and perform a registration estimate every $n$ keyframes as seen in Fig. \ref{fig:FactorGraph}. We optimize the series of poses denoted as:
\begin{equation}
    \mathcal{X}_t = \{P_0, P_1, ..., P_t\}
\end{equation}
\begin{align}
\mathcal{X}_t^* &= \text{argmin}_{\mathcal{X}_t} ||x_o||^2_{\Sigma_0} \nonumber \\
&\quad+ \sum_i^t (||P(x_{i-1},x_i)||^2_{\Sigma_T}) \nonumber \\
&\quad+ \sum_{j}^{t/n} (||R(x_j)||^2_{\Sigma_r})
\end{align}
The initial pose factor is initially the last received GPS measurement to better constrain the optimization. Additionally, we utilize the sensors available aboard the UGV, utilizing Tartan VO\cite{tartanvo2020corl} relative pose factors to reflect the local motion of the robot. We utilize a semi-open loop system, only using GPS readings as a prior estimate for the current at a rate of 0.1Hz to analyze how our system deals with intermittent GPS signal, otherwise, the robot uses the previous optimization result as the initial guess for the pose estimate. Additionally, the registration estimate is used so long as the match is deemed to not be an outlier by the matching module. We add our method's registration estimate at a rate of 5Hz. We solve the optimization with the Levenberg–Marquardt algorithm \cite{levenberg}.

\section{Results}

\subsection{Dataset}
\begin{figure}[!htb]
    \includegraphics[width=\linewidth]{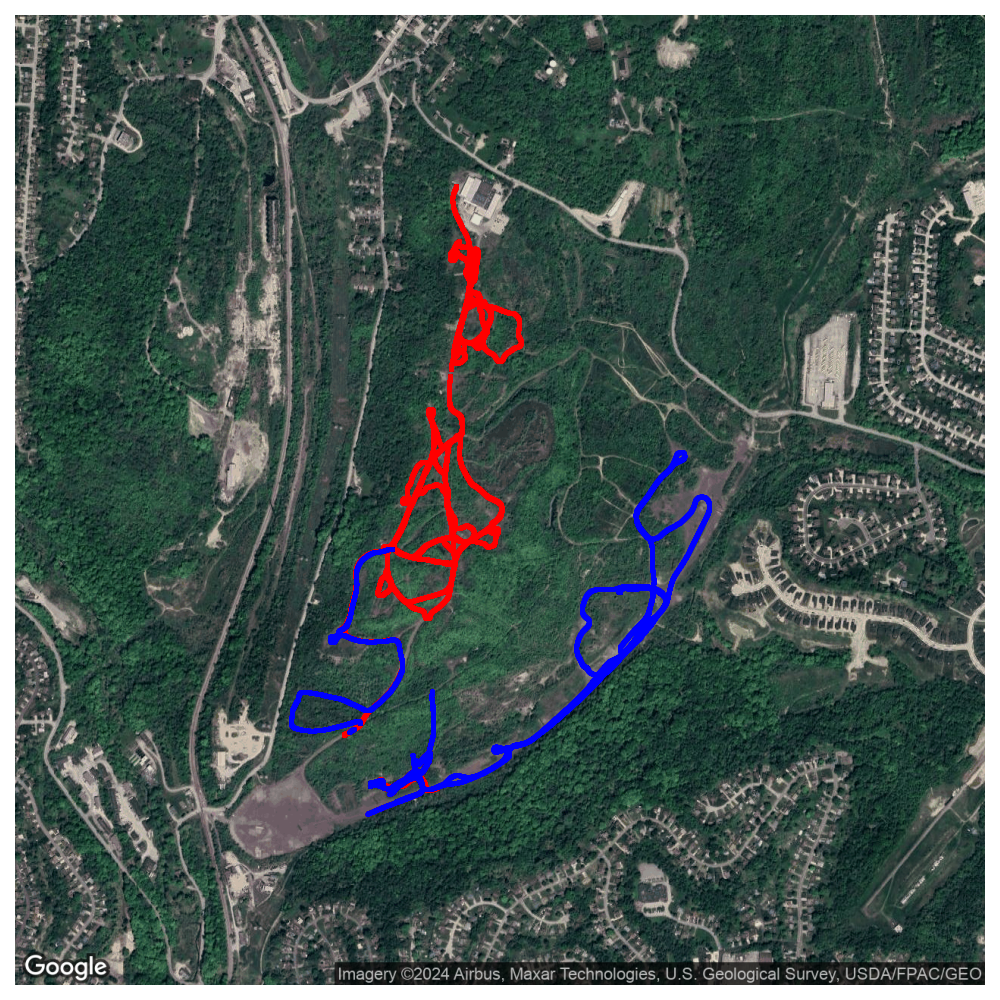}
    \caption{Training and testing splits on TartanDrive 2.0 traced in Google Maps. Training trajectories are in red and testing trajectories are in blue with less than five percent trajectory overlap.}
    \label{fig:MapDatasetSplits}
\end{figure}

For all of our experiments, we utilize the Tartan Drive 2.0 real world dataset which lauds over seven hours of on-road data with images, depth, RGB BEV images, IMU data, and much more \cite{triest2022_tartandrive}\cite{sivaprakasam_tartandrive_2024}. The data rates for images are 10Hz, GPS at 50 Hz, and IMU at 100 Hz. In line with training our network, we preprocess color images and depth images to be the same size and scale the intrinsics. We also consider time synchronization of the different sensors to validate correct operation. For our experiments, we select 15 trajectories for training, 2 for validation, and 3 for testing. To emphasize the generalization capability of our method, we highlight the trajectories for the training and testing splits in Fig. \ref{fig:MapDatasetSplits}

\subsection{Metrics}

\subsubsection{Coarse Matching}
For coarse matching, we analyze the recall, we extract a subset of the map around the last estimated location of size $[S_{crop}, S_{crop}]$ into disjoint and equally sized square map cells of size $[S_{cell},S_{cell}]$. In our work, $S_{crop} = 378$ and $S_{cell} = 32$. For this calculation, we use the ground truth GPS to accurately evaluate how strong the coarse matching is.

We define the size of the BEV local map and size of the grid cell to be the same to simplify the matching process. We define the predicted location as the most probable $k$ grid cells. A  true positive is denoted as one of the $k$ predictions landing in the same grid cell as the ground truth and a false positive for landing in any other cell. We analyze this metric from a granularity from $k = 1$ to $k = 10$. These coarse matches act as a downstream prior for our localization refinement. This metric is crucial as it provides a strong prior for the robot's localization.

\subsubsection{Fine Matching and Trajectory Error}
We calculate the RMSE match from the ground truth position while running as a semi-open loop system. As noted earlier, this semi-open loop system models an intermittent GPS signal at a low frequency. We analyze the relative pose errors given this intermittent GPS signal. This allows us to take into account the quality of the global state estimate towards the optimization and also how good the fine matching is performing individually.

\subsection{Experimental Results}\label{ExperimentalResults}

For coarse localization, we see substantially better results in using the larger foundational models for coarse localization. Our best result is the utilization of the DINOv2-G backbone for the aerial encoder with the ResNet-101 backbone for the ground encoder. It is clear from our results that the more powerful semantic capability from the foundational models reflects improved performance, even though ResNet feature maps have a higher resolution. Fine matching
performs well when coarse estimates provide a strong prior for matching. In line with expectations, ResNet performs strongest at top 1 recall but dips in performance once
more matches are considered. This is likely because ResNet has higher-resolution feature maps, making geometric matching easier but provides less adept at semantic matching. We see a direct correlation between the localization error and the
recall for the given backbone. A visualization of our matching results is present in Fig \ref{Coarse To Fine Matching}

Overall, our method corrects trajectories without reliable GPS signals. The trajectory error error how significant improvement over VO alone. In most cases, the formulation results in the robot staying within the bounds of the trail, using the global state estimate to re-localize when drifting occurs. Although not quite as smooth as the reference trajectory with GPS state estimates, there is still a general success in the resulting localization as in \ref{fig:Traj}. Interestingly, the ResNet-101 model minimizes the trajectory
error and the fine-matching error. This speaks to the value of higher-resolution feature maps for fine-matching tasks, even if performance is diminished for coarse matching.

One failure case involves the registration of global state estimates with low covariance but matches against a hard positive. One example of this is similar road structures in unstructured environments. Also, difficult cases where the map and the ground images differ in semantics are  challenging. One example encountered in the data was the presence of shadows in the aerial image, making matching extremely unlikely. Situations such as these suggest the need for a multi-modal approach to build robustness in the system. We see the introduction of IMU data, range data, and additional outlier filtering as a necessary step to remedy these issues. We leave further optimization, matching, and outlier rejection for future work.

\begin{table}[!htb]
\caption{Recall for Coarse Localization on TartanDrive 2.0 Test Set}
\begin{tabularx}{\linewidth}{|X|l:l:l:l|}
    \toprule
    \textbf{Method} & \textbf{R@1} & \textbf{R@3} & \textbf{R@5} & \textbf{R@10}\\
    \midrule
    RGB Local Map Registration\cite{litman_gps-denied_2022}  & 8.25 & 13.58 & 18.35 & 25.61\\
    BEVLoc ResNet-101 &  \textbf{14.08}& 29.21 & 41.07 & 59.78\\
    BEVLoc DINOv2-B & 13.91 & 31.41 & 44.18 & 61.07\\
    BEVLoc DINOv2-L & 12.77 & 29.29 & \textbf{45.91} & 61.32 \\
    BEVLoc DINOv2-G & 13.48 & \textbf{32.93} & 45.66 & \textbf{65.26}\\
    \hline
\end{tabularx}
\label{tab:Recall}
\end{table}

\begin{table}[!htb]
\caption{Error (meters) for Fine Match and Pose Error on TartanDrive 2.0 Test Set - Figure 8 Sequence}
\begin{tabularx}{\linewidth}{|X|l:l|}
    \toprule
    \textbf{Method} & \textbf{RMSE Match} & \textbf{RPE}\\
    \midrule
    TVO \cite{tartanvo2020corl}& - &90.45\\
    BEVLoc ResNet-101 & \textbf{59.80}  & \textbf{28.23} \\
    BEVLoc DINOv2-B & 72.35 & 37.72 \\
    BEVLoc DINOv2-L & 66.82 & 30.38 \\
    BEVLoc DINOv2-G & 69.36 & 31.33\\
    \hline
\end{tabularx}
\label{tab:Traj}
\end{table}

\section{Conclusion and Future Work}

In our work, we presented a comprehensive framework for deriving global state estimates solely from vision sensors. By leveraging contrastive learning and employing rigorous methods for mining hard negatives, we have successfully learned representations between sequences of ground images and aerial maps. Our devised procedure for coarse-tofine matching utilizing the learned embeddings has shown promising results in unstructured environments, even with intermittent GPS signal availability. 

While our approach demonstrates advancements, it is not without limitations. Challenges such as lighting variations, seasonal changes, and map updates persist, potentially impacting the robustness of our framework. While our framework provides a firm foundation, there is ample room for improvement. 

Future advancements must aim to enhance geometric constraints to improve match quality and reduce reliance on GPS data. Furthermore, utilization of state-of-the-art attention-based matching or other learning techniques can further enhance the capability and robustness of the solution. Additionally, a refinement of alternate outlier rejection methods for registration state estimates and optimized marginalization strategies for non-linear optimization could increase the stability and robustness of the state estimations. By building upon our initial solution and addressing these challenges, we remain optimistic about the potential for significant advancements in the effectiveness and applicability of our framework for future research


\begin{figure}
    \centering
    \includegraphics[width=\linewidth]{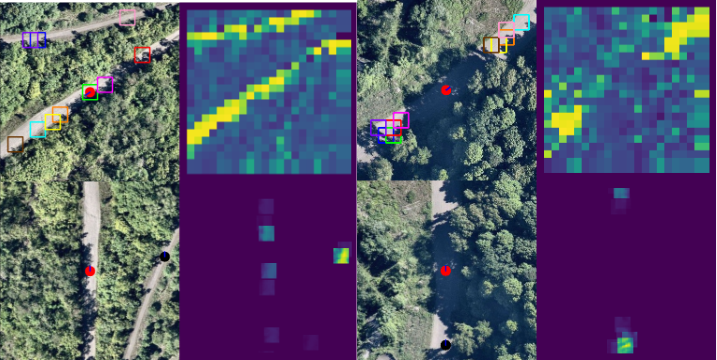}
    \caption{Some failure cases encountered in coarse matching. The left exhibits similar road structure resulting in ambiguous matching. The right shows the difficulty of matching the ground image against shadows.}
    \label{fig:FailureCases}
\end{figure}

\section*{ACKNOWLEDGMENT}
This material is based upon work supported by the U.S. Army Research
Office and the U.S. Army Futures Command under Contract No. W911NF20-D-0002. The content of the information does not necessarily reflect the
position or the policy of the government and no official endorsement should
be inferred.

The authors thank Wenshan Wang, Matthew Sivaprakasam, Lihong Jin, \& Daniel McGann for
their support with the deployment of BEVLoc.

\printbibliography

\end{document}